%% file: main.tex
\documentclass[
]{ceurart}

\usepackage{listings}
\usepackage{graphicx} 
\input{command}

\begin{document}



\copyrightyear{2026}
\copyrightclause{Copyright for this paper by its authors.
  This author version is made available under the Creative Commons
  Attribution 4.0 International License (CC BY 4.0).}

\conference{Preprint of a paper accepted at the 5th Workshop on
LLM-Integrated Knowledge Graph Generation from Text (TEXT2KG),
co-located with ESWC 2026, May 10--14, 2026, Dubrovnik, Croatia}

\title{Leveraging Graph Structure in Seq2Seq Models for Knowledge Graph Link Prediction}

\author[1, 2]{Luu Huu Phuc}[email=huuphuc.luu@de.bosch.com]
\author[1]{Ratan Bahadur Thapa}[email=ratan.thapa@ki.uni-stuttgart.de]
\author[1]{Mojtaba Nayyeri}[email=mojtaba.nayyeri@ki.uni-stuttgart.de]
\author[1]{Jingcheng Wu}[email=jingcheng.wu@ki.uni-stuttgart.de]
\author[2]{Evgeny Kharlamov}[email=evgeny.kharlamov@de.bosch.com]
\author[1, 3]{Steffen Staab}[email=steffen.staab@ki.uni-stuttgart.de]

\address[1]{Analytic Computing, KI, University of Stuttgart, Stuttgart, Germany}
\address[2]{Bosch Center for Artificial Intelligence, Stuttgart, Germany}
\address[3]{WAIS, University of Southampton, United Kingdom}

\begin{abstract}
  We introduce Graph-Augmented Sequence-to-Sequence (GA-S2S), a novel framework that integrates a T5-small encoder-decoder with a Relational Graph Attention Network (RGAT) to improve link prediction in knowledge graphs. While existing Seq2Seq models rely solely on surface-level textual descriptions of entities and relations and at best, flatten the neighborhoods of a query entity into a single linear sequence, thereby discarding the inherent graph structure, GA-S2S jointly encodes both textual features and the full $k$-hop subgraph topology surrounding the query entity. By integrating raw encoder outputs with RGAT's relation-aware embeddings, our model captures and leverages richer multi-hop relational patterns and textual information. Our preliminary experiments on the CoDEx dataset demonstrate that GA-S2S outperforms competitive Seq2Seq-based baseline models, achieving up to a 19\% relative gain in link prediction accuracy.
\end{abstract}


\maketitle

\section{Introduction}
\label{introduction}
Knowledge graphs (KGs) use graph-based data models to represent structured knowledge about real-world entities (e.g., people, places, or things) and their connectedness. They store facts about entities as triples $(e,r,e')$, where $e$ and $e'$ denote entities, and $r$ represents their connectedness, also known as a relation. Reasoning over these structured facts is known to be useful for many artificial intelligence applications, ranging from question answering and recommendation systems to natural language processing~\cite{DBLP:conf/aaai/ZhouHMSZWNS26,DBLP:conf/emnlp/DingWW0XT24}, and so on.

However, the structured facts stored in knowledge graphs in real-world applications often represent only a partial view of the world. This incompleteness arises from practical constraints such as limitations in data acquisition, privacy concerns, contractual restrictions on third-party data sharing, noisy or inaccurate observations (\eg, due to sensor, IoT, or edge device failures), and errors introduced during (human or procedural) data integration~\cite{DBLP:conf/emnlp/ZhuWWZCKS25}. 
As a result, knowledge graphs, whether newly constructed from such data or pre-existing, inherently contain partial information about the world we want to represent, and it is therefore expected that many factual assertions about real-world entities may be either incomplete or entirely absent.
For instance, 71\% of person entities in Freebase lack a recorded place of birth, and 75\% lack a recorded nationality \cite{dong2014knowledge}; moreover, between 69\% and 99\% of entities in YAGO, DBpedia, and similar KGs are missing at least one property that other entities within the same class possess \cite{razniewski2016but}. Link prediction for KGs \cite{kazemi2018simple}, a problem that is early studied under the umbrella of \emph{statistical relational learning} \cite{getoor2007introduction}, aims to infer these missing facts by predicting whether a triple is likely to be true, particularly, by estimating the likelihood of missing entities for a given incomplete triple assertion (\ie, triple query). Formally, given an incomplete triple assertion $(e,r,?)$ or $(?,r,e)$, the objective is to predict the missing entity $e'$ such that the newly formulated triple $(e,r,e')$ or $(e',r,e)$ corresponds to a plausible missing fact in the KGs.

Conventional link prediction methods for KGs can be classified into \emph{embedding-based} approach \cite{kg_emb_analysis_paper} and \emph{Graph Neural Network} (GNN)-based approach \cite{gnn_for_kg_survey_paper}. 
Embedding-based methods assign distinct vectors to each entity and relation, employing scoring functions to evaluate triple plausibility \cite{transE-paper,rescal-paper,complex-paper}. Their parameter sizes scale linearly with the number of entities and relations, making them memory-intensive to train on large-scale KGs. Moreover, by treating each triple in isolation, static embeddings often fail to capture higher-order or multi-hop structures, thereby limiting their ability to generalize over broader subgraph patterns. GNN-based methods, on the other hand, address this limitation by leveraging the graph structure through iterative neighborhood message passing and aggregation \cite{rgcn-paper, rgat-paper,compgcn-paper}. Nonetheless, neither embedding-based nor GNN-based approaches are primarily designed to process unstructured information (\ie, textual descriptions, literal attributes, \etc) without integrating specialized or auxiliary modules \cite{text-enhanced-KG-embedding-survey}.
Both embedding and GNN approaches typically answer a query, for example, a head-relation query $(e,r,?)$, by computing a score $f(e,r,e')$ for every tail entity candidate $e'$ in the knowledge graph (KG) and then ranking them to select the top-$k$, a process that is $O(|E|)$ per query - where $|E|$ is the number of entities in the KG - and becomes computationally expensive for large KGs (\ie, $|E|\geq$ millions). 

To address the computationally expensive inference for large KGs, recent generative approaches leveraging transformer-based Sequence-to-Sequence (Seq2Seq) models, such as KGT5 \cite{kgt5}, KG-S2S \cite{kg-s2s-paper}, KGT5-context \cite{kgt5-context}, and UniLP \cite{uniLP-paper}, have achieved state-of-the-art results on standard link-prediction benchmarks, such as Wikidata5M \cite{kepler-paper} and WikiKG90Mv2 \cite{ogb-dataset-paper}. These methods, by contrast to the embedding and GNN-based approaches, assume that entities and relations can be verbalized with unique text mentions and recast link prediction as a text-generation task. For example, given the verbalized forms ``\tx{Michael Jackson}'', ``\tx{occupation}'', and ``\tx{musician}'' for a triple, the query $(e_{\text{Michael~Jackson}}, r_{\text{occupation}}, ?)$ is reformulated as input sequence ``\tx{Predict tail: Michael Jackson} $\vert$ \tx{occupation}'', and the Seq2Seq model is then trained to generate the target mention ``\tx{musician}''. Since Seq2Seq-based methods employ a fixed parameter budget independent of the number of entities and relations, and replace the scoring-ranking step by directly generating the text mentions of target entities, they overcome the linear parameter growth and high inference cost in previous approaches.
However, since these models operate solely on verbalized text mentions, 
they do not explicitly model topological patterns (\eg, multi-hop neighborhood), which trades off the possibility of exploiting subgraph structural information in KGs.

In this paper, we address this limitation of the Seq2Seq-based approach by proposing a hybrid model that integrates a Seq2Seq backbone based on the T5-small model \cite{t5-paper}, initialized without pre-trained weights, with a relation-aware GNN module, \ie, Relational Graph Attention Network (RGAT) \cite{rgat-paper}. Rather than constraining the input to a single flat sequence, our model is able to process the neighborhood topological structure of a query entity. The proposed model encodes each query and its full $k$-hop neighborhood into a shared latent space using the T5 encoder. The resulting representations are then enriched via relation-aware message passing over the neighborhood surrounding the query entity, using RGAT. Finally, the augmented representation is decoded by the T5-decoder to generate the verbalized text mention of the target entity. This design simultaneously leverages textual information as well as local graph structures. We benchmark our approach on the CoDEX dataset, which contains 3 KGs ranging from tens of thousands to hundreds of thousands of triples, and show that it outperforms competitive Seq2Seq-based baseline methods in link prediction accuracy.

\section{Background}
\label{session:background_problem_setting}

A knowledge graph $G$ is a (multi)set of triples, each drawn from the Cartesian product $E \times R \times E$, where $E$ and $R$ are two disjoint sets of entities and relations. Each triple $(h,r,t) \in G$ represents the fact that a relation $r \in R$ exists between head entity $h \in E$ and tail entity $t \in E$. 

Generative Se2Seq-based methods excel at leveraging textual data for link prediction~\cite{DBLP:journals/tmlr/DingL0NWTBM25}. They reformulate queries such as $(e, r, ?)$ as text-generation tasks, \ie, generating the text mention $s(e')$ for a target entity $e'$ that best completes the query triple $(e, r, e')$. It is therefore assumed that all entities $e$ and all relations $r$ in KGs are associated with a unique text mention, denoted by $s(e)$ and $s(r)$, respectively. That is, for any $e,e'\in E$ and $r,r'\in R$, we have $e=e'\leftrightarrow s(e)=s(e')$ and $r = r'\leftrightarrow s(r) = s(r')$.

KGT5 \cite{kgt5} is one of the pioneering works to cast knowledge graph link prediction as a Seq2Seq task, translating pairs of input queries and target entities into text sequences and training a T5-small model from scratch to predict the missing entity. Building on this, KG-S2S \cite{kg-s2s-paper} unifies static, temporal, and inductive link prediction under a single Seq2Seq paradigm. KGT5-context \cite{kgt5-context} further enhances KGT5 by prepending linearized 1-hop neighbor triples of the query entity to the input sequence. For example, an input might read:
``\tx{Predict tail: Michael Jackson} $\vert$ \tx{occupation} $\vert$ \tx{Context: genre $\vert$ pop music $\vert$ record label $\vert$ Sony Music $\vert$ \dots}'',
where the ``Context'' segment verbalizes all $1$-hop triples linked to ``Michael Jackson''. Going beyond 1-hop triples, UniLP \cite{uniLP-paper} introduces soft prompts, learnable embeddings representing 1-hop relations, into each transformer layer. This approach allows the model to implicitly capture structural information, improving its ability to encode the topology of the knowledge graph without relying solely on textual context.

\section{The GA-S2S Model}
\label{section:proposed_model}
\begin{figure}[htbp]
\centering
\includegraphics[width=0.95\linewidth]{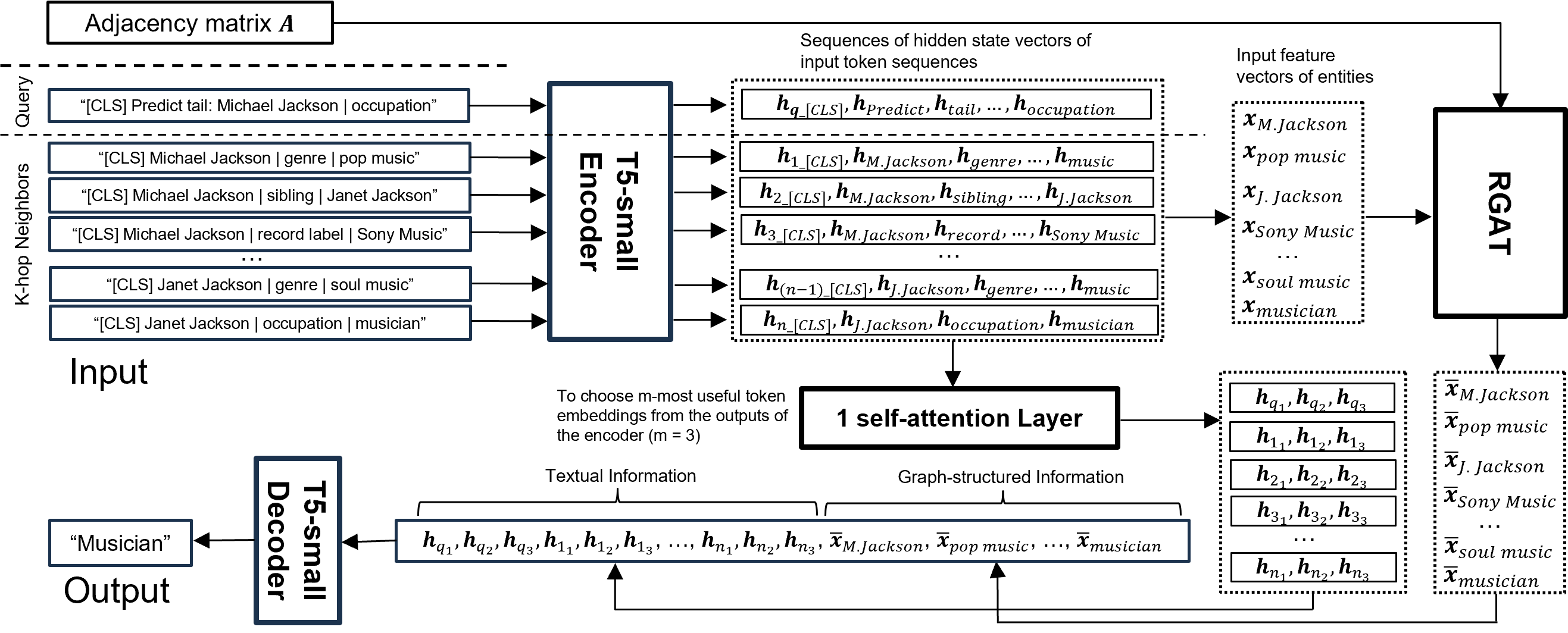}
\caption{The proposed model architecture integrates an RGAT module between the encoder and decoder of the T5-small model, enabling the fusion of textual and structural information from KGs. All components of the T5 model and the RGAT module are fully trainable and initialized from scratch, and no pre-trained weights are frozen during training.}
\label{fig:model_architecture}
\end{figure}

The GA-S2S model, as shown in Figure \ref{fig:model_architecture}, is composed of 3 sequential components: a text encoder, an RGAT-based GNN module, and a text decoder. We employ the encoder and decoder from the T5-small model without pre-trained weights. We initialize the T5 model from scratch because pre-trained weights are heavily optimized for natural language syntax. Given that our inputs are highly structured, template-based verbalizations mixed with graph embeddings, training from scratch prevents potential negative transfer from natural language priors.

\paragraph{Input Representation.}
Given a query $q=(e, r, ?)$, the input to our model consists of the text sequence $s(q)$ corresponding to the query, a set of text sequences $s( G_{e,k})$  of a sampled $k$-hop subgraph $ G_{e,k}$ of the query entity $e$, and the associated adjacency matrix $\mathbf{A}$ of $ G_{e,k}$.

We first sample a $k$-hop subgraph 
 $G_{e,k}=\{(h,r,t)\in G\mid h\in E_{e,k} \wedge t\in E_{e,k}\}$
 of the query entity $e$ and integer $k\geq 0$, where $ E_{e,k} =\{e'\mid \text{ there exists a path of length $\leq k$ between $e$ and $e'$ in } G\}$. Each triple in the induced subgraph $ G_{e,k}\subseteq G$ of the original $G$ under the query $q$ is then verbalized as, for any $i\leq n=|G_{e,k}|=|\{(h,r,t)\in G\mid h\in E_{e,k} \wedge t\in E_{e,k}\}|$, 
 \[s(h_i,r_i,t_i)= \text{``[CLS]''} s(h_i) \text{``}\mid\text{''} s(r_i) \text{``}\mid\text{''} s(t_i),\]
 where ``[CLS]'' denotes a special token of the Seq2Seq model.
 The connectivity of $ G_{e,k}$, i.e., its adjacency matrix $\mathbf{A}$, is encoded via 2 tensors: a $n\times 2$ tensor of node-index pairs $[[h_1, t_1], \dots, [h_n, t_n]]$, and a length-$n$ tensor of relation indices $[r_1, \dots, r_n]$.

\paragraph{Encoding and Subgraph Representation.}
All text sequences $s(q)\cup s(G_{e,k})$ are fed to the T5-encoder, producing a sequence of hidden state vectors $H=\{ \mathbf{h}_{\tx{tok}_1}, \dots, \mathbf{h}_{\tx{tok}_l}\}$ for each input text sequence, where $l\leq 512$. Since each entity $e'\in E_{e,k}$ in the sampled subgraph $G_{e,k}$ can appear in one or more triples, we aggregate its representation by averaging over the $\mathbf{h}_\tx{[CLS]}$ vectors of all triples it participates in, denoted by $\mathbf{x}_{e'}$:
\[
\mathbf{x}_{e'}  = \frac{1}{|G_{e',k}|} \sum_{ (h,r,t)\in G_{e', k}} \mathbf{h}_{[\tx{CLS}]}^{(h,r,t)},
\]
where $G_{e',k} = \{(h,r,t) \in G_{e,k} \mid h = e'\vee t = e'\}$ and $\mathbf{h}_{[\tx{CLS}]}^{(h,r,t)}$ is the hidden state vector of ``[CLS]'' token of $s(h,r,t)$. We then pass the set of entity features $\{\mathbf{x}_{e'}\}$ and adjacency matrix $\mathbf{A}$ into the RGAT module, which propagates and aggregates information across the subgraph, yielding updated embeddings $\{\overline{\mathbf{x}}_{e'}\}$.

\paragraph{Decoding with Combined Representations.}
We add a skip connection from the encoder to the decoder so that the decoder receives both the raw encoder output and graph-enhanced embeddings from the GNN module. 
For each input sequence $s \in \{s(q)\} \cup s(G_{e,k})$, the encoder produces a corresponding sequence of token-level hidden state vectors $H$.
We apply a lightweight self-attention layer to distill $H$ into the $m$ most salient vectors $\{\mathbf{h}_1,\dots,\mathbf{h}_m\}$, as not all tokens in $s$ contain useful information. 
Empirically, we find that setting $m=3$ provides a good balance between performance and efficiency. 
The decoder input is the concatenation of these distilled vectors for the query and its $k$-hop neighborhood triples with the RGAT’s output node features $\{\overline{\mathbf{x}}_{e'}\mid e'\in E_{e,k}\}$.
 
The entire pipeline is trained end-to-end by minimizing the cross-entropy loss between the decoder’s generated token distributions and the ground-truth entity mentions.

\section{Experimental Settings}
\label{session:experiments}
We use CoDEx dataset \cite{codex-paper} in our experiments, which is extracted from Wikidata and Wikipedia. Unlike earlier benchmarks, CoDEx is particularly designed to mitigate train/test leakage and avoid trivial relational patterns. CoDEx also provides multilingual textual labels for entities and relations; in our experiments, we use English textual labels as the text mentions. Detailed statistics of the dataset are shown in Table~\ref{tab:codex_dataset}.

\begin{table}[htbp]
    \centering
    \caption{Statistics of the 3 KGs in CoDEx dataset}
    \label{tab:codex_dataset}
    \begin{tabular}{lccc}
        \toprule
         & CoDEx-S & CoDEx-M & CoDEx-L  \\
        \midrule
        \#Entity & 2034 & 17050 & 77951  \\
        \#Relation & 42 & 51 & 69  \\
        Train & 32888 & 185584 & 551193  \\
        Valid & 1827 & 10310 & 30622 \\
        Test & 1828 & 10311 & 30622 \\
        \midrule
        Avg Deg & 32.3 & 21.7 & 14.1 \\
        Median Deg & 17 & 12 & 7  \\
        Density & 0.0159 &  0.0012 &  0.0002 \\
        \bottomrule
    \end{tabular}
\end{table}

We train our model with a batch size of 64 and set all hidden dimensions to 512. We tune the RGAT dropout rate over $\{0.0, 0.1, 0.2, 0.5\}$ and the number of attention heads over $\{1, 2, 4\}$ via grid search. Our RGAT implementation is based on \emph{PyTorch Geometric} \cite{pytorch-geometric}, and we use the T5-small encoder, decoder, and transformer layers from \emph{HuggingFace} \cite{huggingface}. For each query, we sample a 1-hop neighborhood of size 75 for CoDEx-L, and for CoDEx-S/M, we use neighborhood sizes of 75 for 1-hop and (10, 5) for 2-hop neighborhoods. We restrict to 2-hop neighborhoods only for the smaller KGs due to the computational limitations of our current model implementation.

We evaluate using Mean Reciprocal Rank (MRR) and Hit@k $(k=1, 3, 10)$, following the standard scoring and ranking procedure described in \cite{kgt5}. For baseline comparison, we evaluate against the current state-of-the-art generative models for KG link prediction, namely KGT5 \cite{kgt5}, KG-S2S \cite{kg-s2s-paper}, and KGT5-context \cite{kgt5-context}. UniLP \cite{uniLP-paper} is not included in our evaluation due to the unavailability of its codebase.

\section{Overall Evaluation}
\label{session:evaluation}
Table~\ref{tab:codex_lp_results} shows link prediction performance on the CoDEx dataset. At first glance, GA-S2S outperforms all other baselines on CoDEx-M/L and ranks second on CoDEx-S. GA-S2S with 1-hop neighborhoods improves the MRR scores over KGT5-context (also using 1-hop neighborhoods) by 11.2\%, 4.8\%, and 1.7\% on CoDEx-S/M/L respectively. Incorporating 2-hop neighborhoods further increases MRR by 19.5\% on CoDEx-S and 6.6\% on CoDEX-M. While existing Seq2Seq methods are constrained to 1-hop (flattened) neighborhoods, these results suggest that $k$-hop neighborhoods with $k \geq 2$ are advantageous for the task. The improvement is most noticeable on CoDEx-S and decreases as the KG size increases. This is expected since the bigger KGs are much sparser with lower degrees of connectivity as shown in Table~\ref{tab:codex_dataset}.

\begin{table}[htbp]
    \centering
    \caption{Comparison of link prediction performance on CoDEx dataset. The best results are written in bold. The second-best results are underlined. All results are averaged over 4 runs. Comparison is done using a t-test with p-value = $0.01$.}
    \label{tab:codex_lp_results}
    \resizebox{\linewidth}{!}{%
    \begin{tabular}{lcccccccccccc}
        \toprule
         &
        \multicolumn{4}{c}{CoDEx-S} &
        \multicolumn{4}{c}{CoDEx-M} &
        \multicolumn{4}{c}{CoDEx-L} \\
        \cmidrule(lr){2-5} \cmidrule(lr){6-9} \cmidrule(lr){10-13}
        & MRR & H@1 & H@3 & H@10 & MRR & H@1 & H@3 & H@10 & MRR & H@1 & H@3 & H@10 \\
        \midrule
        KG-S2S & 0.269 & 0.182 & 0.314 & 0.454 & 0.240 & 0.175 & 0.267 & 0.376 & 0.235 & 0.181 & 0.260 & 0.343 \\
        KGT5 & \textbf{0.344} & \textbf{0.269} & \textbf{0.387} & \underline{0.492} & 0.276 & 0.214 & 0.314 & 0.398 & 0.267 & 0.212 & 0.299 & 0.375 \\
        KGT5-context & 0.277 & 0.186 & 0.313 & 0.471 & 0.271 & 0.206 & 0.304 & 0.402 & \underline{0.292} & \underline{0.236} & \underline{0.321} & \underline{0.400} \\
        GA-S2S (1-hop) & 0.308 & 0.210 & 0.359 & \textbf{0.512} & \underline{0.284} & \underline{0.219} & \underline{0.317} & \underline{0.415} & \textbf{0.297} & \textbf{0.241} & \textbf{0.327} & \textbf{0.404} \\
        GA-S2S (2-hop) & \underline{0.331} & \underline{0.252} & \underline{0.381} & 0.489 & \textbf{0.289} & \textbf{0.223} & \textbf{0.322} & \textbf{0.421} & - & - & - & - \\
        \bottomrule
    \end{tabular}%
    }
\end{table}

Although KGT5-context extends KGT5 with 1-hop neighborhood information, its performance is lower than that of KGT5 on the CoDEX-S/M datasets. This suggests that naively appending linearized 1-hop neighborhoods of query entities to input sequences can negatively impact model performance.

It is equally noteworthy that KGT5, even without any explicit neighborhood graph structure input, remains the top performer on CoDEx-S. We hypothesize that this is due to the excessively long and potentially redundant decoder input sequences in both our model and KGT5-context. Even when limiting each neighbor triple to three embedding vectors, large neighborhoods can still result in unwieldy decoder inputs in GA-S2S. 
As discussed above, flattening all information into one unstructured sequence may confuse the decoder and impede correct generation. A decoder redesign that accepts structured input could therefore yield further performance gains.

\vspace{-0.11cm}
\section{Limitations}
\label{section:limitations}

While Se2Seq models often outperform embedding methods on large-scale KGs with millions of entities \cite{kgt5,kgt5-context}, they do not consistently exhibit a performance advantage over embedding methods on small- to mid-sized KGs.
This is reflected in the CoDEx benchmark: GA-S2S, along with other existing Seq2Seq-based models, lags behind leading KG embedding methods.  For instance, ComplEx \cite{complex-paper} achieves MRR scores of up to 0.465 and 0.337 on CoDEx-S and CoDEx-M, respectively, and TuckER \cite{TuckER-paper} achieves an MRR score of 0.309 on CoDEx-L. Nevertheless, the performance gap between GA-S2S and embedding-based methods narrows as the KG size increases from CoDEx-S to CoDEx-L.

Additionally, GA-S2S currently incurs higher computational costs than standard Seq2Seq models and faces scalability challenges on very large KGs. This overhead arises from the richer processing of neighborhood structures via the GNN module, which leads to a higher number of processed input tokens per query (up to $512 \times$ neighborhood size for GA-S2S versus only up to $512$ for baselines). However, this complexity directly contributes to the model’s ability to jointly capture graph topology with textual information. Thus, our preliminary exploration with the GNN module establishes a basic hybrid Seq2Seq architecture for further improvement. In particular, incorporating graph transformers \cite{graphformer-paper,knowformer-paper}, whose designs are more compatible with transformer-based encoders and decoders, might offer a promising direction. By natively processing graph-structured data via topology-aware attention mechanisms, graph transformers can inherently avoid the need to flatten neighborhoods into linear sequences, thereby enabling deeper integration of structural and textual data.

\section{Conclusion and Open Questions}
\label{session:conclusion_future_work}

We have introduced GA-S2S, a hybrid Seq2Seq-GNN framework that directly integrates graph structure into generative Seq2Seq link prediction models. In contrast to prior Seq2Seq methods that simply linearize neighborhood triples into input sequences, GA-S2S employs a relation-aware GNN (RGAT) to encode both local neighborhood triples and $k$-hop subgraph topology, enabling richer multi-hop relational reasoning. On the CoDEx benchmarks, our approach consistently outperforms competitive Seq2Seq baselines, particularly when incorporating 2-hop neighborhood, demonstrating the clear advantages of explicitly encoding graph structure. 


Despite these promising results, several open questions remain for advancing hybrid Seq2Seq models, as highlighted in Sect. \ref{section:limitations}. First, how can more balanced and rigorously evaluated neighborhood sampling strategies be designed to manage computational overhead without sacrificing structural context? Second, could efficient encoder-decoder architectures, such as hierarchical transformers \cite{hierarchical-transformer}, serve as a more scalable backbone for massive KGs? Finally, what are the optimal techniques for seamlessly integrating textual and structural information throughout both the encoding and decoding stages? Addressing these questions will be critical for the next generation of knowledge graph reasoning models.

\begin{acknowledgments}
The authors thank the International Max Planck Research School for Intelligent Systems (IMPRS-IS) for supporting Jingcheng Wu. Jingcheng Wu and Ratan Bahadur Thapa has been funded by the Deutsche Forschungsgemeinschaft (DFG, German Research Foundation) - SFB 1574 - Project number 471687386.
\end{acknowledgments}

\section*{Declaration on Generative AI}
During the preparation of this work, the authors used Gemini in order to: Grammar and spelling check. Specifically, GenAI tools were used only for minor improvements to the authors' written text. After using these tools, the authors reviewed and edited the content as needed and take(s) full responsibility for the publication’s content.

\bibliography{main}

\end{document}

%% file: command.tex
\usepackage{graphicx} 
\usepackage{subfiles}
\usepackage{centernot}
\usepackage{stackengine}
\usepackage{gensymb}
\usepackage{hyperref}
\usepackage{xspace}

\newcommand*\ie{\text{i.e.,}\xspace}

\newcommand*\eg{\text{e.g.,}\xspace}

\newcommand*\etc{\textit{etc}\xspace}
\newcommand\tx[1]{\texttt{#1}}